\pdfoutput=1

\documentclass[11pt]{article}

 \usepackage[final]{acl}
\usepackage{times}
\usepackage{latexsym}

\usepackage[T1]{fontenc}

\usepackage[utf8]{inputenc}

\usepackage{microtype}

\usepackage{inconsolata}

\usepackage{graphicx}
\usepackage{caption}
\usepackage{subcaption}
\usepackage{hyperref}
\usepackage{soul}

\usepackage{booktabs}
\usepackage{multirow}
\usepackage{amsmath}
\usepackage{amsfonts}

\usepackage{enumitem}

\title{Analogical Structure, Minimal Contextual Cues and Contrastive Distractors: Input Design for Sample-Efficient Linguistic Rule Induction}

\author{
 \textbf{Chunyang Jiang\textsuperscript{1,2}},
 \textbf{Paola Merlo\textsuperscript{1,2}}
\\
\\
 \textsuperscript{1}Idiap Research Institute, Switzerland
 \textsuperscript{2}University of Geneva, Switzerland
\\
 \texttt{Firstname.Lastname@unige.ch}
}

\begin{document}
\maketitle

\begin{abstract}
Large language models achieve strong performance on many tasks, but their training makes it hard to see which properties of the input support efficient linguistic rule learning.
We ask how three cognitively-inspired principles of input design support sample-efficient linguistic rule induction:  analogical structure, contrastive learning, and minimal contextual cue. We also ask how their effects compare to those of LLMs on the same controlled tasks.
We implement these principles in structured sentence completion tasks that test English verb alternations.
Lightweight models trained on hundreds to one-thousand such examples learn the alternation rules with high F1 on these tasks.  
Ablation studies show that analogical organisation is the main driver of sample efficiency, and contrastive distractors and minimal context help further gains. 
We also evaluate zero- and few-shot LLMs on the same tasks. In this controlled setting, the lightweight models reach higher F1 with far fewer task-specific data.
We treat this contrast as a comparison between learning regimes rather than a general verdict on LLMs.
Our results show that careful input organisation supports sample-efficient learning of linguistic rules and reveals distinct learning signatures for trained lightweight models and prompted LLMs.

\end{abstract}

\section{Introduction}

Analogical reasoning, recognising parallelism of structural relationships across different contexts, enables efficient learning by abstracting patterns from minimal examples and transferring knowledge to novel situations~\citep{lake2017building}. However, this principle has faced persistent scalability challenges across computational approaches: early symbolic systems required extensive knowledge engineering~\citep{forbus1995mac}, neural word embeddings showed inconsistent performance~\citep{mikolov2013efficient}, and even transformer-based models like BERT show mixed capabilities~\citep{ushio-etal-2021-bert, thrush2020investigating}. 
While recent systems achieve strong performance~\citep{yasunaga2024large,jiayang-etal-2023-storyanalogy, wijesiriwardene-etal-2023-analogical}, they still require substantial computational resources or show inconsistent results across complexity levels.

Rather than scaling analogical reasoning systems, we operationalise analogical principles through strategic input organisation. 
Research shows that processing constraints can optimise attention allocation~\citep{christiansen2016now}, while contrastive learning frameworks show how systematic positive-negative comparisons improve discriminative learning~\citep{chen2020simple, he2020momentum}.
This suggests that structural organisation, not architectural scaling, may unlock analogical efficiency.

\begin{figure*}[t!]
\centering
\includegraphics[width=\textwidth]{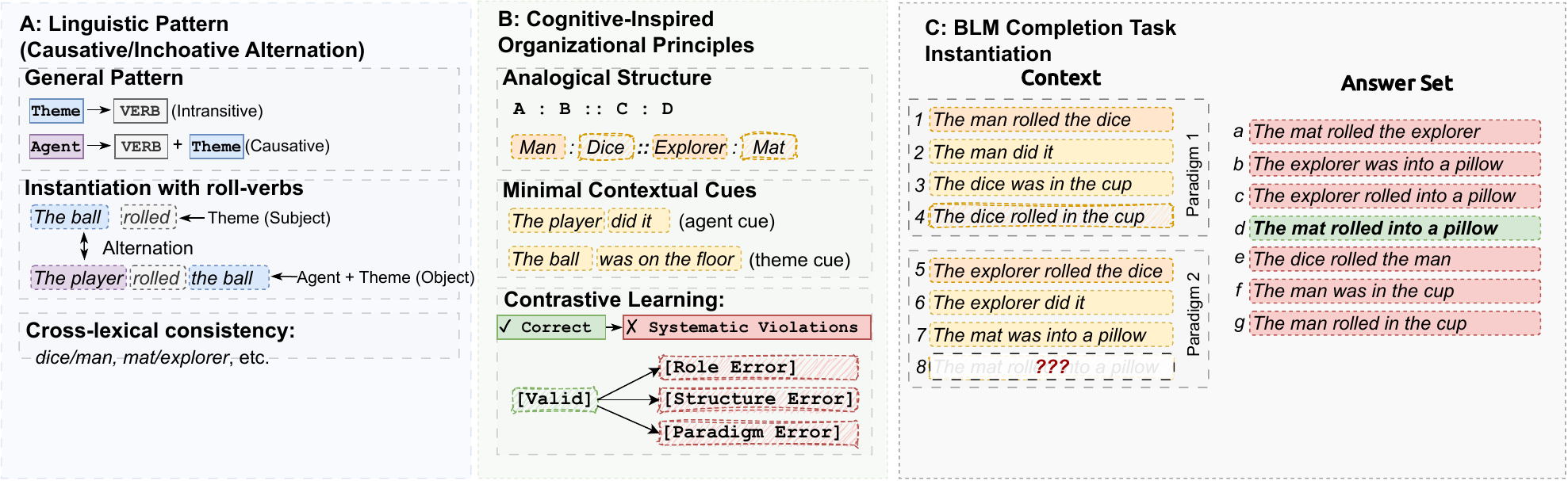}
\caption{Analogical paradigm organisation for sample-efficient linguistic rule learning. (A)~Causative/inchoative alternation pattern showing systematic Agent$\leftrightarrow$Theme mapping in English \textit{roll}-class verbs, with cross-lexical consistency across multiple verb instances. (B)~Three cognitive-inspired organisational principles: analogical structure enables cross-paradigm pattern recognition (\textit{Man:Dice :: Explorer:Mat}), contrastive learning provides discriminative boundaries through systematic constraint violations, and minimal contextual cues offer semantic scaffolding without explicit labelling. (C)~Implementation through structured completion tasks where models must integrate all three principles to identify correct answer \textbf{d} from systematically designed alternatives, each testing specific aspects of analogical reasoning capability. (Distractor taxonomy details in Table~\ref{tab:data-roll-error-def}).}
\label{fig:framework_overview}
\end{figure*}

In this paper, we ask how input organisation can support sample-efficient linguistic rule induction. We also ask how the behaviour of lightweight models that use such organisation compares to that of large language models on the same controlled tasks.
We develop a computational approach that organises input data into analogical paradigms through three cognitively-inspired strategies: (i) \textit{Analogical Structure}, the  systematic paradigmatic relationships that support pattern recognition; (ii) \textit{Contrastive Learning}, the systematically designed distractors that enable discriminative learning; and (iii) \textit{Minimal Contextual Cues}, the implicit annotations providing essential semantic information without explicit labelling.

We evaluate this setup using English verb alternations~\citep{Levin93} through structured completion tasks based on Blackbird Language Matrices~\citep{merlo-2023-blackbird}.
These tasks require identifying correct sentence completions from systematically organised contexts with contrastive alternatives. 
Figure~\ref{fig:framework_overview} illustrates our approach: contexts encode analogical mappings across parallel paradigms (e.g., \textit{Man:Dice :: Exploring:Mat}), requiring models to abstract structural and semantic role relationships rather than memorise surface forms.

Results demonstrate sample efficiency: lightweight models (BERT+CNN, $\sim0.5M$ parameters) trained on only $100$ structured examples achieve $F1=0.95$,  and, in this setting, is higher than zero-shot \texttt{GPT-o3} ($F1=0.87$) on the same tasks. Cross-phenomenon validation confirms robustness beyond causative constructions.

This paper makes three contributions. First, we introduce a computational approach that organises input into analogical paradigms for sample-efficient linguistic rule induction. Second, we provide empirical evidence that this approach enables competitive performance for lightweight models with minimal training examples. Third, we show that, in this controlled setting, lightweight models reach higher F1 than zero-and few-shot LLMs on the same tasks, under a different learning regime.

In all experiments, lightweight models are fine-tuned on $100-1 000$ task-specific examples. LLMs are evaluated in zero-shot and few-shot prompt-based settings without parameter updates. We therefore treat this as a comparison between learning regimes rather than a general verdict on LLM capabilities.

\section{Data and Task Design}
\label{sec:data}

Our data consist of structured sentence completion tasks based on Blackbird Language Matrices~~\citep{merlo-2023-blackbird}. We extend this framework to instantiate three cognitively inspired principles of input design.
This section describes our task framework, linguistic domain, and distractor design methodology.

\subsection{Task Framework}
Our completion tasks organise linguistic data into analogical paradigm structures. 
Each instance contains a structured~\textit{Context} comprising two parallel paradigms encoding analogical relationships (Pattern A $\leftrightarrow$ Pattern B), supplemented with minimal contextual cues that provide semantic scaffolding without explicit labelling. It also contains a systematic \textit{Answer Set}: One correct completion that maintains established patterns, and six systematically designed distractors that violate specific structural, semantic, or paradigmatic constraints.

Success requires integrating our three cognitive principles. Models must recognise analogical mappings across paradigms (analogical structure), discriminate between valid and invalid transformations (contrastive learning), and use semantic scaffolding from minimal soft annotations (minimal contextual cues).

  \begin{table*}[ht!]
\centering
\small
\begin{tabular}{llcccl}
\hline
\textbf{Type} & \textbf{Example} & \textbf{P} & \textbf{S} & \textbf{R} & \textbf{Tests} \\
\hline
Correct & \textit{The mat rolled into a pillow} &$\times$&$\times$&$\times$& All principles (multi-layered analogical reasoning) \\
\hline
RR & \textit{The explorer rolled into a pillow} &$\times$&$\times$& $\checkmark$ & Semantic roles (minimal contextual cues) \\
SC-RR & \textit{The explorer was into a pillow} &$\times$& $\checkmark$ & $\checkmark$ & Syntactic structure (contrastive learning) \\
SCRS & \textit{The mat rolled the explorer} &$\times$& $\checkmark$ & $\checkmark$ & Argument structure (syntax + semantics) \\
\hline
PC-RR & \textit{The man rolled in the cup} & $\checkmark$ &$\times$& $\checkmark$ & Cross-paradigm consistency (analogical structure) \\
PSC-RR & \textit{The man was in the cup} & $\checkmark$ & $\checkmark$ & $\checkmark$ & Multiple constraints (analogical + contrastive) \\
PSC-RS & \textit{The dice rolled the man} & $\checkmark$ & $\checkmark$ & $\checkmark$ & Complex analogical mapping (all principles) \\
\hline
\multicolumn{6}{l}{\footnotesize P=Paradigm, S=Structure, R=Role violations} \\
\end{tabular}
\caption{Systematic distractor taxonomy testing cognitive principles through hierarchical constraint violations. Each error type violates specific linguistic dimensions ($\checkmark$) to isolate analogical reasoning components.}
\label{tab:data-roll-error-def}
\end{table*}

\subsection{Linguistic Test Domain: English Verb Alternations}
\paragraph{Causative/Inchoative Alternations}
We use English verbs in the \textit{roll}-class ~\citep{Levin93}, such as \textit{roll}, \textit{drop} and \textit{slide}. These verbs participate in the causative/inchoative alternation: they appear both in an intransitive frame (\textit{The ball rolled}) and a transitive causative frame (\textit{The player rolled the ball}). This alternation requires systematic Agent$\leftrightarrow$Theme mappings, where the same entity can take different grammatical roles depending on causation.

This linguistic phenomenon provides an ideal test case because it requires analogical reasoning across different lexical items sharing the same structural pattern. It also needs to discriminate between valid and invalid argument structure transformations, and integrate semantic role information with syntactic patterns.

\paragraph{Cross-Phenomenon Validation}
We validate the generalisability of the approach with \textit{bake}-class verbs, a class of verbs exhibiting the unspecified object alternations (\textit{The chef baked a cake} vs. \textit{The chef baked}). In this class, the subject of both the transitive and the intransitive is an \textit{Agent}. While the two classes share syntactic structure, they instantiate semantically distinct processes, testing whether our organisational principles generalise beyond specific semantic classes.

\subsection{Analogical Paradigm Organisation}

Figure~\ref{fig:framework_overview} illustrates our systematic organisation approach. Each context follows a $2 \times 4$ paradigmatic structure.

\paragraph{Multi-Layered Analogical Structure}
Contexts encode analogical relationships at multiple levels: \textit{Man:Dice :: Explorer:Mat} in terms of Agent-Theme relationships, with identical structural transformations applied across different lexical content. This tests whether models can abstract relational patterns beyond surface similarity.

\paragraph{Minimal Contextual Cues}
We provide semantic scaffolding through subtle annotations within each paradigm.
For example, action descriptors indicate agentivity (\textit{The player did it}) while state descriptions suggest thematic roles (\textit{The ball was on the floor}).
These cues clarify semantic relationships without explicit grammatical labelling.

\paragraph{Contrastive Learning Through Systematic Distractors}
Each answer set implements contrastive learning by providing systematic negative examples alongside each correct completion.
Table~\ref{tab:data-roll-error-def} defines our hierarchical error taxonomy designed to test specific cognitive principles through constraint violations.

Each distractor violates exactly one constraint dimension while preserving others, enabling precise evaluation of rule component acquisition. Role errors test semantic understanding, structural errors test syntactic knowledge, and paradigm errors test analogical consistency across contexts.

Notice that distractors are contextually inappropriate rather than inherently ungrammatical. For example, \textit{The explorer rolled into a pillow} is grammatical but violates the analogical mapping established in the context.

\subsection{Data Generation}
Our systematic generation process creates sentences based on templates with controlled lexical variation and seed alternation pairs, following established verb classifications~\citep{Levin93}.\footnote{Verb class and seed sentences are detailed in Appendix~\ref{sec:app:ling_phenomenon}.}
We follow a hybrid, expert-guided process. Linguistic experts select and validate seed sentences and design a small set of templates with soft annotations for each alternation class. We then automatically instantiate full paradigms from these seeds and templates, controlling lexical variation and error types.
This procedure creates a dataset of structured completion tasks over \textit{roll}- and \textit{bake}-class verbs, which we refer to as \textsc{BLM-Roll-BakeE}.
\footnote{Data is available at \href{https://www.idiap.ch/en/scientific-research/data/blm-roll-bakee/}{https://www.idiap.ch/en/scientific-research/data/blm-roll-bakee/}}

\section{Experiments} \label{sec:exp}

We systematically evaluate our analogical paradigm organisation approach through controlled experiments testing three core claims: (i) our organisational principles enable sample efficiency compared to unstructured baselines; (ii) individual components contribute systematically to performance; (iii) lightweight models trained with structured data can match zero-shot and few-shot LLMs on the same controlled linguistic rule learning tasks, under a different learning regime.
The performance on BLM puzzle completion is used as a proxy for a model's level of successful linguistic rule learning. 

\subsection{Experimental Design}
\paragraph{Ablation Framework:} We test five systematic conditions that isolate organisational components while controlling information content. 

\textsc{Base} is the complete paradigmatic organisation with analogical structure and contextual cues. 
\textsc{Shuffled} has identical content in random order, (tests organisational vs. content effects). 
\textsc{NoAnalogy} removes the first paradigm (tests analogical structure contribution).
\textsc{NoSoftCue} removes contextual annotations (tests scaffolding contribution). 
\textsc{Transposed} preserves analogical patterns but changes spatial arrangement (
\( C_{\text{Trans}} = C_{\text{Base}}^\top \)).
\footnote{See the formal definitions in Appendix~\ref{sec:app:ablation}.}

\paragraph{Data Configurations:}
\textit{Type I} uses the same verbs across paradigms ($3 000$ examples, 80:10:10 splits, direct analogical mapping). 
\textit{Type II} uses different verbs across paradigms ($15 000$ examples, same splits, tests abstraction). 
\textit{Cross-phenomenon} uses unspecified object alternations with \textit{bake}-class verbs (validates generalisability).

\paragraph{Scale Testing}
Training sizes range from $10$ to $2 700$ examples for lightweight models; and $300$-example test sets (same as for \textit{Type I} test data, and $300$ random examples from \textit{Type II} test set) for LLMs.
This setup balances computational feasibility with statistical reliability for cross-model comparisons.

\subsection{Models and Training}
\label{sec:architecture}

\paragraph{Lightweight Models}
We use several models of representation. We embed each sentence using
BERT\footnotemark \citep{devlin-etal-2019-bert} as our primary model, \footnotetext{\texttt{bert-base-multilingual-cased}} 
 RoBERTa\footnotemark ~\citep{liu-etal-2019-roberta} and 
 \footnotetext{\texttt{xlm-roberta-base}}
 ELECTRA\footnotemark~\citep{clark-etal-2020-electra} for encoder comparison.
 We reason that using a weaker base clarifies contributions attributable to data organisation rather than encoder capabilities.
 \footnotetext{\texttt{electra-base-discriminator}}

As for architecture types, we use previously tested parameter-efficient neural models (CNN and FFNN in \citet{nastase-merlo-2023-grammatical} for comparison). The CNN captures localised patterns and positional relationships, while the  FFNN processes concatenated embeddings through fully-connected layers, integrating distributed information across the entire context. Both architectures maintain minimal parameter counts ($\sim0.5M$ parameters) compared to the encoder ($\sim110M$), isolating the influence of data organisation from architectural capacity.
Both models output an answer embedding given the concatenated embeddings of the sentences from the context, evaluated against answer candidates using a max-margin objective with cosine similarity. Training runs for $120$ epochs, with learning rate $0.001$, Adam Optimizer, batch size $100$, and early stopping (patience $10$). All experiments use random seed $42$ with results averaged over $3$ independent runs.\footnote{Full encoder comparison and replication details are provided in Appendix~\ref{sec:app:model_spec}.}

\begin{figure}[h!]
\includegraphics[width=0.9\linewidth]{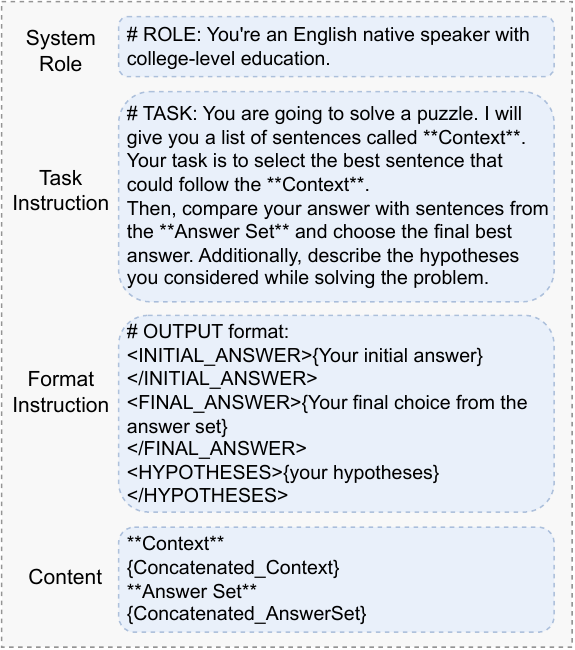}
\caption{Zero-shot prompt.}
    \label{fig:prompt}
\end{figure}

\paragraph{Large Language Models}
We evaluate eight advanced models including reasoning-capable systems.

\textit{Reasoning models}: \texttt{deepseek-R1}~\citep{deepseekai2025deepseekr1incentivizingreasoningcapability}, \texttt{gpt-o3}, \texttt{gpt-o3-mini},\texttt{qwq-32B}.

\textit{Standard models}: \texttt{deepseek-V3}~\citep{deepseekai2024deepseekv3technicalreport}, \texttt{llama-3.3-70B-Instruct}, \texttt{llama-3.2-3B-Instruct}, \texttt{qwen3-32B}.

The $temperature$ is $0.1$, where applicable (standard models), and the $max\_tokens$ is $2046$.
\footnote{Details on LLM configurations are in Appendix~\ref{sec:app:llm_config}.}

\begin{figure}[h!]
\centering
\includegraphics[width=0.9\linewidth]{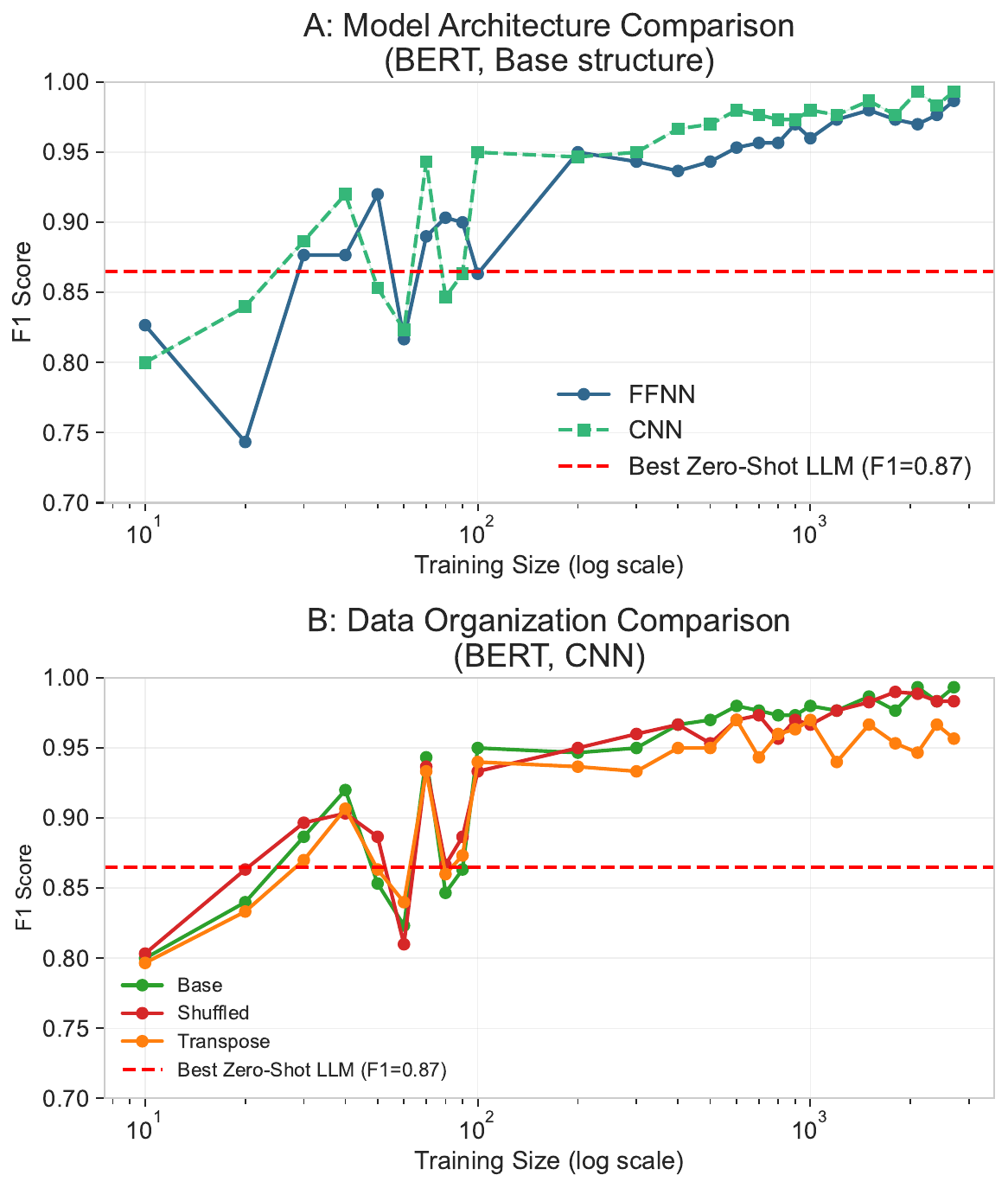}
\caption{F1 Performance as a function of training size. (A) Comparison of model architectures with BERT embeddings on Base structure organisation. (B) Impact of data organisation using the best architecture. }
    \label{fig:exp_f1_f1_size}
\end{figure}

\paragraph{Prompting and Evaluation}
We test zero, one and five-shot prompting with (w-CoT) and without (wo-CoT) chain-of-thought reasoning~\citep{wei2021finetuned, kojima2022large}.
We use a puzzle-solving prompt that parallels our lightweight model setting. Figure~\ref{fig:prompt} shows a zero-shot example.
The LLM first sees the context, generates a provisional answer, then compares it to the answer set before generating the best final sentence (instead of answer key label), and it also generates the hypotheses considered. Any final choice not in the answer set is flagged as a system error (\texttt{ERR}). This prompt mimics our learning objective using pretrained models and presents a more challenging task than standard multiple-choice questions.
Randomising the options in the answer set and asking to generate sentences instead of option labels help debias LLMs' token bias (e.g., A/B/C/D) in multiple-choice-format questions~\citep{zheng2024large}. 
This uniform puzzle-solving format enables direct comparisons across model scales and training paradigms.

\paragraph{Evaluation} F1 scores balance precision and recall across 7 answer choices. 
Responses not matching the answer set options are flagged as system errors for LLMs.
A multi-factor ANOVA tests Structure $\times$ Model $\times$ Shots $\times$ CoT interactions ($p < 0.05$).

\subsection{Results}

\begin{figure}[t!]
    \centering
        \includegraphics[width=0.9\linewidth]{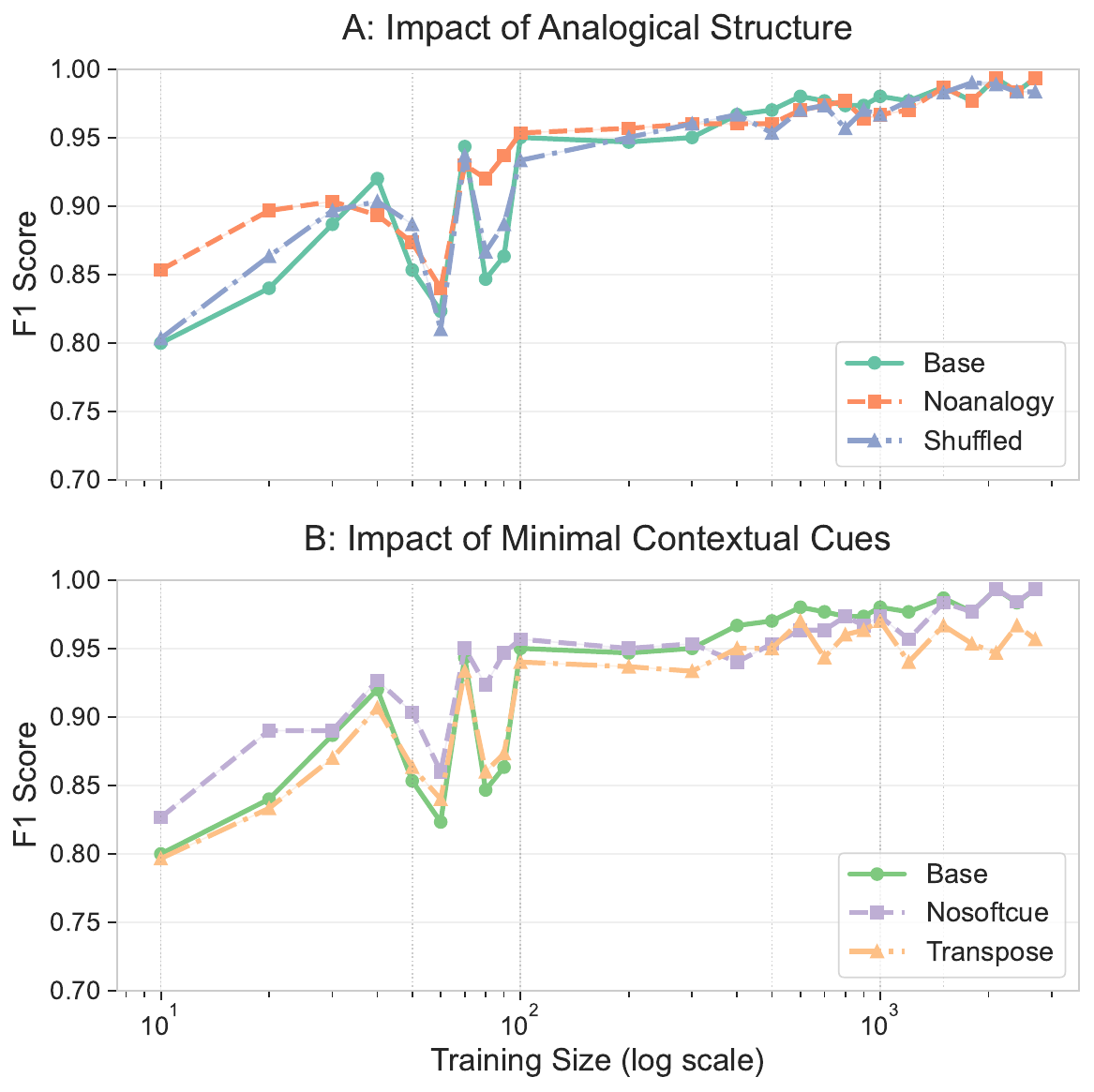}
    \caption{Isolated contributions of organisational components to best model performance. (A)~ Impact of analogical organisation, comparing \textsc{Base} against \textsc{NoAnalogy} and \textsc{Shuffled}. (B)~ Impact of implicit soft annotations, comparing \textsc{Base} with \textsc{NoSoftCue} and \textsc{Transposed} variants.}
    \label{fig:exp_f3_compare_structure}
\end{figure}

\begin{figure*}[ht!]
    \centering
    \includegraphics[width=\linewidth]{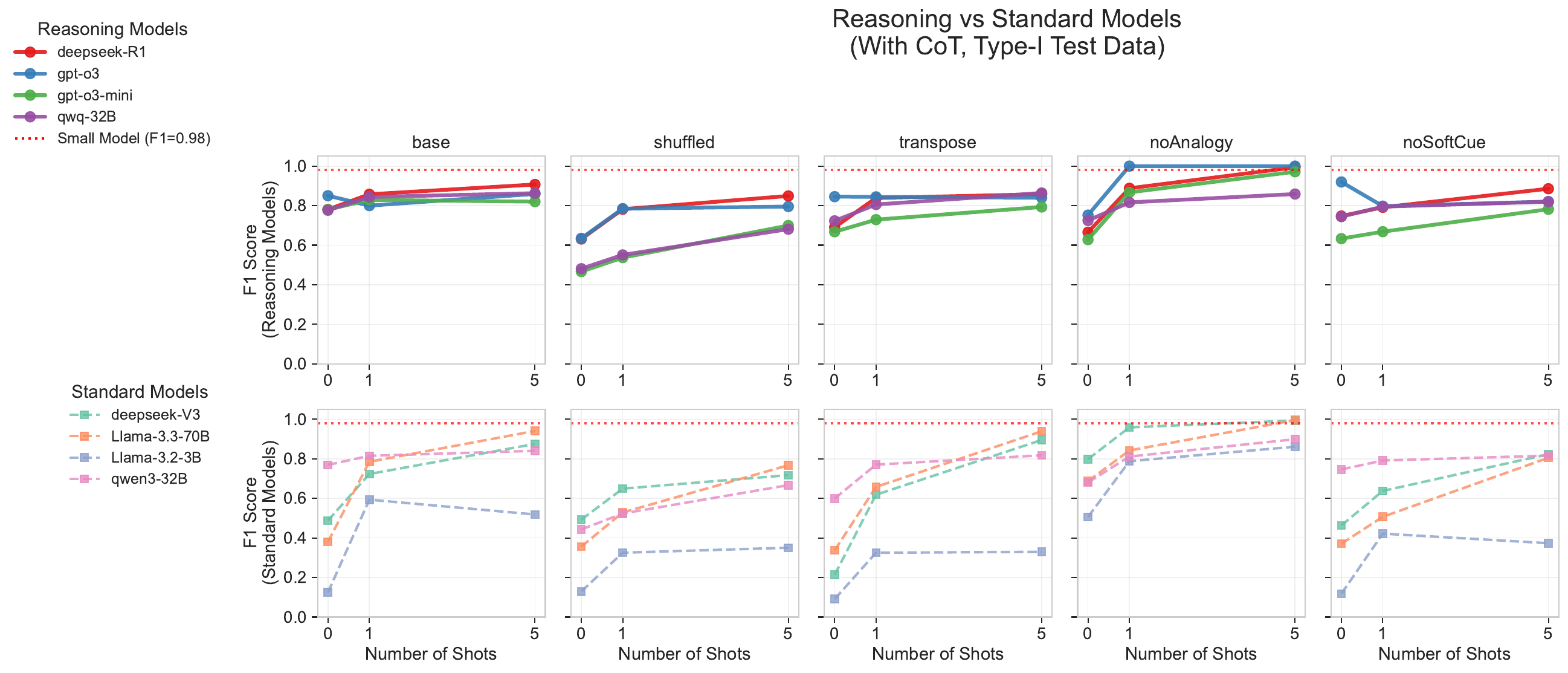}
\caption{Reasoning models (top row: \texttt{deepseek-R1}, \texttt{gpt-o3}, \texttt{gpt-o3-mini}, \texttt{qwq-32B}) vs. standard models (bottom row: \texttt{llama-3.3-70B-Instruct}, \texttt{llama-3.2-3B-Instruct}, \texttt{qwen3-32B}, \texttt{deepseek-V3}) across data structures. Red dotted line shows small model baseline ($F1=0.98$). Reasoning models consistently outperform standard models, but struggle to match structured lightweight model performance in zero-shot settings.}
    \label{fig:f4_llm}
\end{figure*}

We present results from our best-performing architecture (CNN) across all experimental conditions with lightweight models. While both CNN and FFNN demonstrate similar qualitative trends, CNN consistently outperforms FFNN across training regimes, likely due to its capacity to capture local sequential patterns critical for analogical mapping.

\subsubsection{Sample Efficiency Validation} \label{sec:size-var}

Figure~\ref{fig:exp_f1_f1_size} confirms our core finding: lightweight models with \textsc{Base} organisation achieve $F1=0.95$ with only $100$ examples, and, in this setting, are higher than our best zero-shot reasoning model \texttt{GPT-o3} ($F1=0.87$) on the same tasks.
Performance stabilises at $1 000-1 200$ examples across architectures.
Panel B shows the impact of data organisation, where \textsc{Base} structure outperforms \textsc{Shuffled} and \textsc{Transposed} arrangements. 
Identical content produces different learning trajectories, confirming that organisational structure, not information quantity, drives efficiency gains.

\subsubsection{Component Analysis} \label{sec:structure-var}

Figure~\ref{fig:exp_f3_compare_structure} validates the systematic contributions from individual organisational components. While all structures eventually converge to similar performance levels at larger training sizes, analogical structure (\textsc{Base} vs. \textsc{NoAnalogy} vs.\textsc{Shuffled}) shows slightly more consistent advantages in the $500-1 500$ example range where data efficiency matters most.
Contextual cues (\textsc{Base} vs. \textsc{NoSoftCue} vs. \textsc{Transposed}) provide additional but smaller benefits.

These subtle differences in learning curves suggest that while both components contribute to learning, modern neural architectures can compensate for deficiencies in either component as training sizes increases.

\subsubsection{LLM Comparison} \label{sec:llm-compare}

Figure~\ref{fig:f4_llm} confirms our third contribution: lightweight models trained on $1 000$ structured examples reach higher $F1$ than all tested LLMs in zero-shot conditions, including state-of-the-art reasoning models, on these tasks and under our training regime. Multi-factor ANOVA reveals significant main effects for structure, model, and number of shots, with minimal CoT impact\footnote{
Complete results on Multi-factor ANOVA appear in Appendix~\ref{sec:app:llm_anova} (Table~\ref{tab:multi-anova-llm}).
}. This result suggests that this task engages pattern recognition and rule application rather than multi-step reasoning.\footnote{Complete results on LLMs performance are in Appendix~\ref{sec:app:llm_results} (Table~\ref{tab:llm_all}).}

\paragraph{Reasoning vs.\ Standard Models}
Reasoning models start with high zero-shot performance ($F1=0.8+$) but plateau quickly, while standard models begin lower ($F1=0.1-0.5$) but show steeper improvement curves, making them more sensitive to organisational structure.

\paragraph{Organisational Structure Patterns}
\textsc{Shuffled} consistently performs worst across all models, confirming that random organisation disrupts learning. \textsc{Base} and \textsc{NoAnalogy} emerge as most effective, often performing comparably. Notably, \textsc{NoAnalogy} sometimes outperforms \textsc{Base} in zero-shot settings for reasoning models, suggesting analogical structure becomes less crucial when models already possess strong reasoning capabilities. However, standard models benefit most from organisational effects that have greater room for improvement.

\subsection{Error Analysis and Patterns}

\begin{figure*}[ht!]
    \centering
              \includegraphics[width=\linewidth]{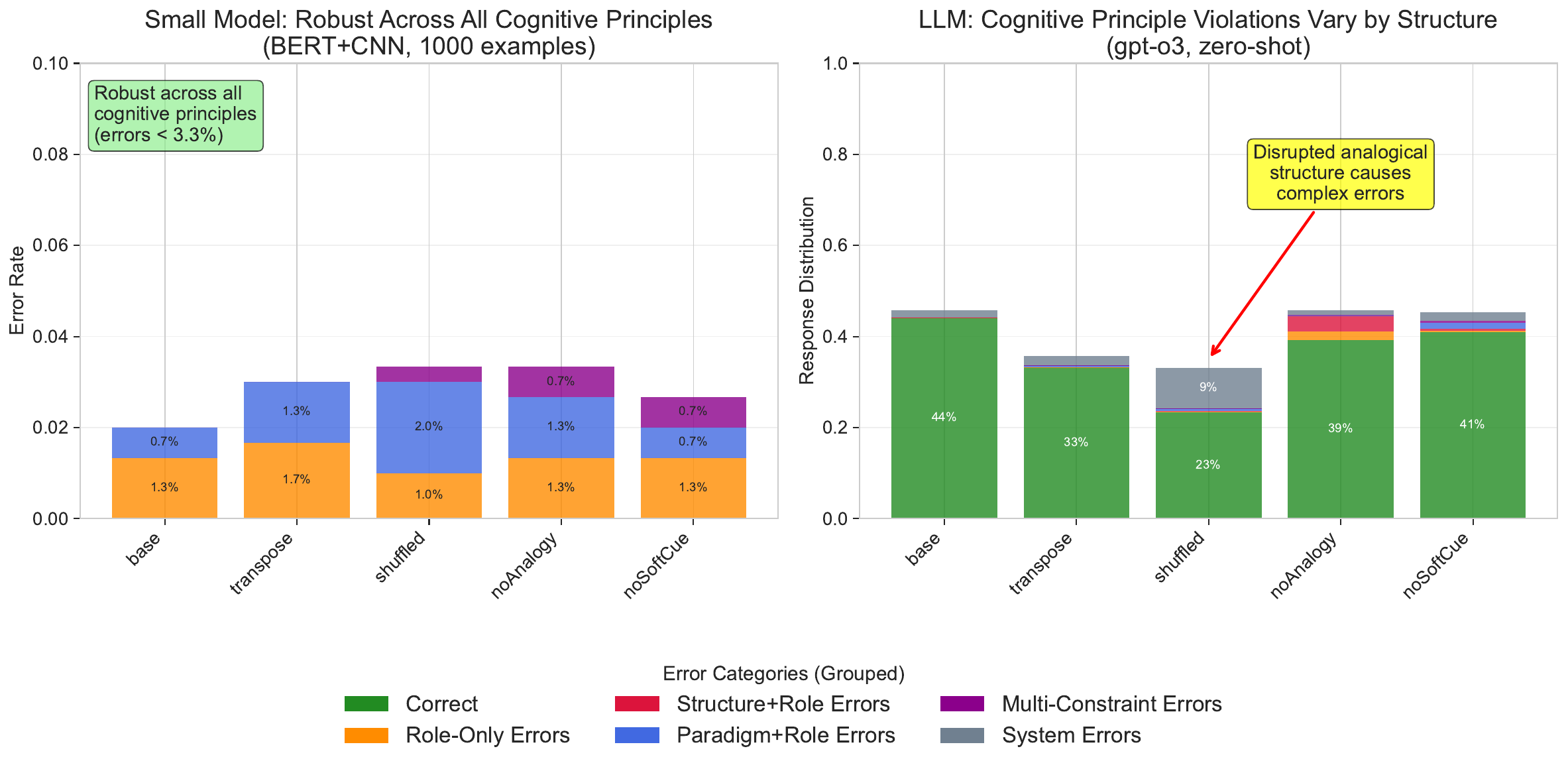}
\caption{Error analysis across data organisations. Lightweight models (left) show robust performance with minimal errors ($<3.3\%$) across all structures. LLMs (right, \texttt{gpt-o3} zero-shot) show structure-dependent error patterns following our cognitive-inspired error taxonomy (Table~\ref{tab:data-roll-error-def}). \textsc{Shuffled} structures particularly disrupt LLMs' analogical reasoning, causing increased multi-constraint violations.}
\label{fig:5a}
\end{figure*}

\begin{figure}[ht!] 
\centering 
\includegraphics[width=\linewidth]{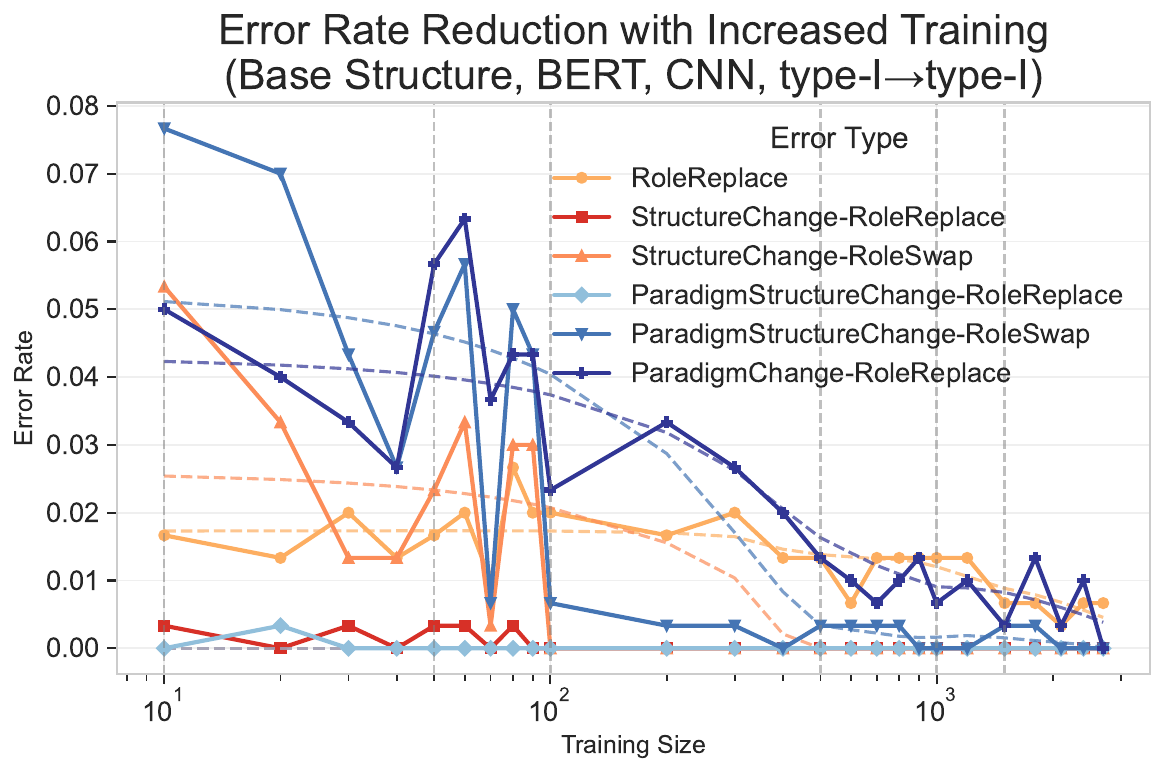} 
\caption{Reduction in error rates as training size increases for the \textsc{Base} structure (train and test on \textit{Type I} data). Dashed lines represent smoothed trends using LOWESS smoothing.} 
\label{fig:f5c-app} 
\end{figure}

Figure~\ref{fig:5a} shows organisational structure affects error patterns across different model types.
Lightweight models are robust ($errors<3.3\%$) across all organisational conditions, demonstrating consistent learning regardless of structural variations.
LLMs show structure-dependent error patterns following our error taxonomy. Disrupted analogical structure (\textsc{Shuffled} condition) causes increased multi-constraint violations, with correct responses dropping from $55\%$ (\textsc{Base}) to $24\%$ (\textsc{Shuffled}) for \texttt{gpt-o3}. This shows that even reasoning-capable models rely heavily on structural organisation for linguistic rule learning tasks.

Figure~\ref{fig:f5c-app} tracks error reduction as training progresses with the \textsc{Base} structure. ParadigmChange-RoleReplace errors show the most dramatic decrease, which suggests that paradigm-level information is learnt earlier than fine-grained role distinctions. The \textsc{Base} structure facilitates more efficient error reduction across all types, which provides better implicit feedback about linguistic rules, consistent with error-driven learning theories.

\subsection{Additional Analysis}

\paragraph{Cross-Phenomenon Generalisation}
\begin{figure}[ht!]
\includegraphics[width=\linewidth]{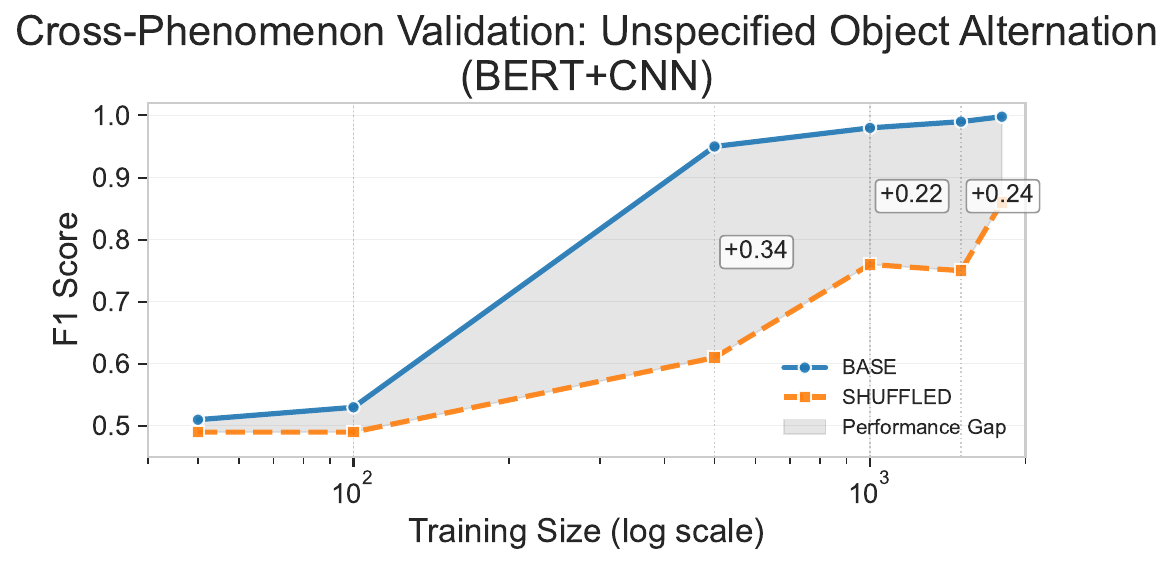}
\caption{Cross-phenomenon validation using unspecified object alternations (\textit{bake}-class verbs). Results show $F1$ performance using BERT+CNN architecture on \textit{Type II} data, averaged over 3 runs.}
\label{fig:cross_bake}
\end{figure}

Figure~\ref{fig:cross_bake} shows that \textsc{Base} results are higher than \textsc{Shuffled} also with \textit{bake}-class verbs.
This suggests that our organisational benefits (\textsc{Base} $>$ \textsc{Shuffled}) generalise beyond causative alternations and indicates that our approach captures general organisational principles rather than construction-specific optimisations.

\begin{figure}[ht!]

\includegraphics[width=\linewidth]{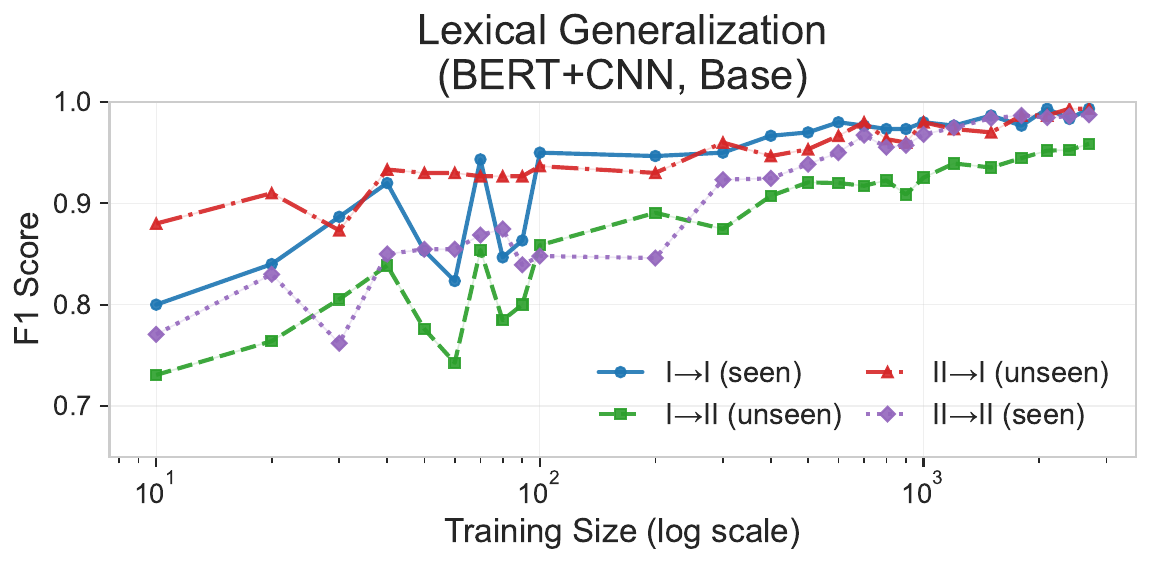}
    \caption{Seen type vs.\ Unseen Type performance across training sizes. 
    }
    \label{fig:exp_f2_lexical_generalisition}
\end{figure}

\paragraph{Cross-Type Generalisation} Figure~\ref{fig:exp_f2_lexical_generalisition} shows successful rule abstraction: models achieve robust cross-type performance by $100-200$ examples, though all conditions eventually converge to high performance with sufficient training data. This confirms that structural relationships rather than surface patterns are being learnt.

\begin{figure}[ht!] 
\centering 
\includegraphics[width=\linewidth]{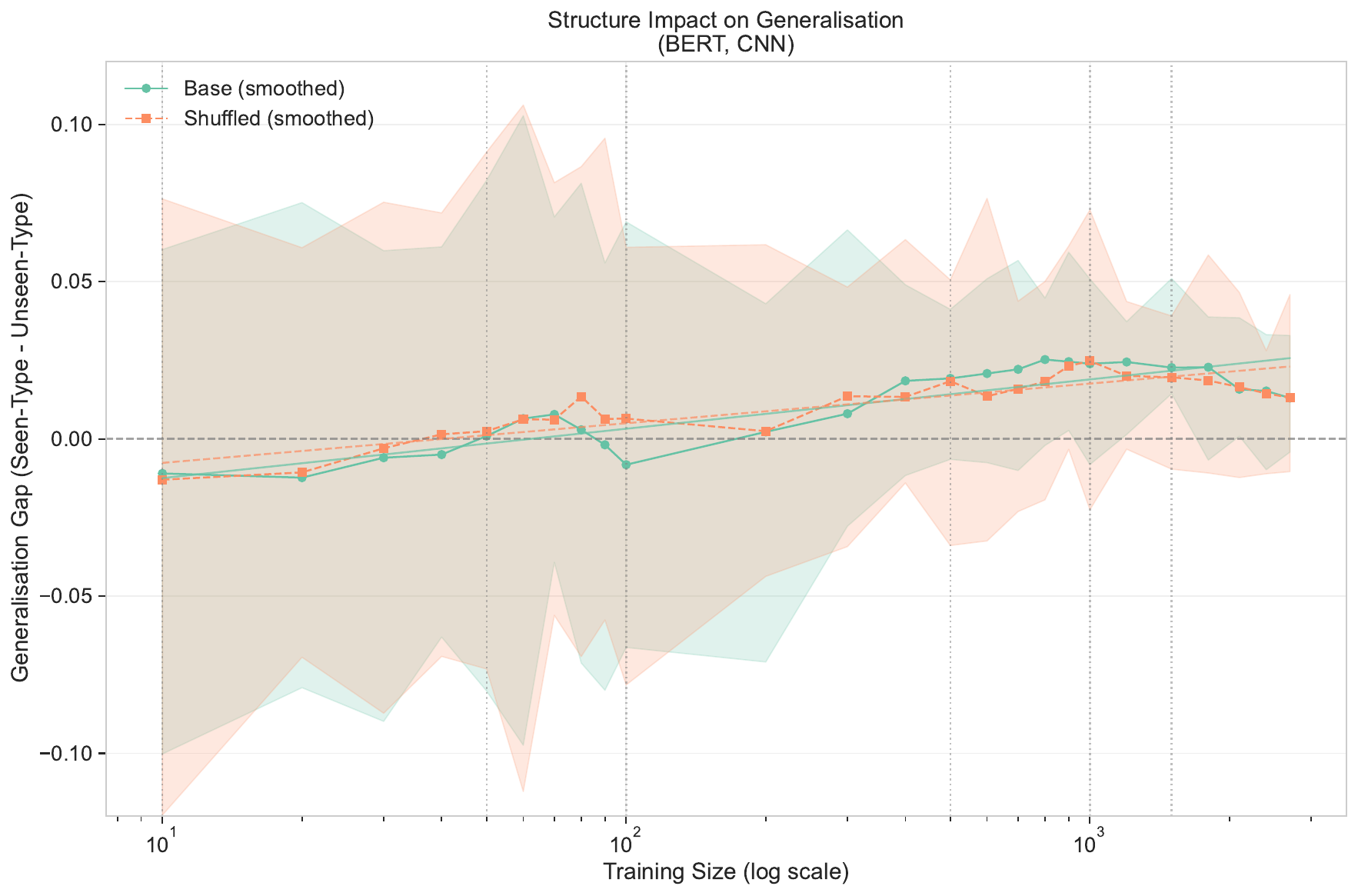} 
\caption{Generalisation gap between \textsc{Base} and \textsc{Shuffled} structures. Smaller gaps indicate better generalisation. The \textsc{Base} structure facilitates better transfer of linguistic rules across training sizes.} 
\label{fig:generalization-app} 
\end{figure}

Figure~\ref{fig:generalization-app} shows that the \textsc{Base} maintains more stable and efficient generalisation throughout the learning process, while the \textsc{Shuffled} shows erratic performance in mid-range training sizes. At small training sizes, both structures show negative gaps, which suggests that the models avoid overfitting to surface lexical patterns.\footnote{Details for the generalisation gap metric and smoothing method appear in Appendix~\ref{sec:app:setup}.}

\paragraph{Discussion}

Our systematic evaluation validates all our three core claims. The finding that lightweight structured models ($F1=0.95$) exceed even \texttt{gpt-o3} in zero-shot settings on our tasks  ($F1=0.87$) demonstrates that systematic input organisation can achieve strong linguistic rule learning on controlled tasks with reduced task-specific data and computational requirements. 
The consistent organisational effects across both reasoning and standard LLMs indicate that analogical paradigm organisation addresses fundamental challenges in pattern recognition that persist even in the most advanced current systems. 
The contrasting error patterns between model types suggest that organisational benefits operate through different mechanisms, with accelerated learning for lightweight models and improved error discrimination for LLMs.
The cross-phenomenon and cross-type evaluation confirms that our approach captures general principles rather than construction-specific optimisations.

These findings indicate that structured input data organisation is complementary to performance gains for linguistic rule learning beyond architectural advances or reasoning capabilities.

\section{Related Work}

\paragraph{Structured Data for Sample Efficiency}
Achieving competitive performance with reduced training needs remains a central challenge in NLP.
Recent work emphasises how reorganising raw text into structured or hierarchically arranged contexts can significantly enhance model learning \citep{liu2024enhancing, sourati2024arn}. 
``Structured data'' often refers to knowledge graphs, table data, databases \citep{jiang-etal-2023-structgpt,li2023resdsql}. \citet{liu2024enhancing} instead reorganise text into hierarchical formats (e.g., Scope-Aspect-Description) to mimic human knowledge consolidation, thereby improving task performance in question-answering. 
Existing sample efficiency approaches in NLP include data selection~\citep{albalak2024a}, data augmentation~\citep{kumar-etal-2020-syntax, dai-etal-2023-auggpt}, curriculum learning~\citep{bengio2009curriculum, xu-etal-2020-curriculum}, among others.
However, they typically focus on scaling or sequencing existing data rather than fundamental reorganisation principles.
Our work instead targets linguistic rule learning through structure in a structured completion tasks inspired by cognitive principles.

\paragraph{Analogical Learning in Neural Language Models} 
Analogical reasoning is the ability to perceive and use similarities on relational patterns rather than surface features~\citep{gentner2012analogical}. 
Neural models have shown strong capabilities of analogical reasoning to generalise rules, such as word-level analogies~\citep{mikolov2013efficient, brown2020language,ushio-etal-2021-bert,webb2023emergent}, concept analogies through analogical structure abduction~\citep{yuan-etal-2023-beneath}, narrative-based or story-level analogies~\citep{nagarajah2022understanding,jiayang-etal-2023-storyanalogy, sourati2024arn}, or analogical prompting where LLMs generate relevant exemplars or knowledge before solving the problem~\citep{yasunaga2024large}. These approaches, like ours, suggest that explicit structural cues improve analogical processing.

In linguistics and cognitive science, work on argument structure and alternations has argued that children track families of related verbs and constructions when they learn causative alternations, rather than memorising isolated patterns. \citet{bowerman1974learning} studies how children acquire English causative verbs and the relation between cognitive, semantic, and syntactic development.
Our approach is computationally inspired by this line of research, but it does not aim to model child learning directly.
We instead study how analogical organisation of input, contrastive distractors, and minimal contextual cues affect sample efficiency and generalisation behaviour of neural models on controlled alternation tasks.

\paragraph{Contrastive Learning in NLP} 
Contrastive learning has emerged as a powerful principle for neural representation learning. Systematic comparison between positive and negative examples enhances discriminative learning~\citep{chen2020simple, he2020momentum}. 
However, existing contrastive approaches typically require large-scale training data and focus on representation learning rather than rule acquisition. 
Our systematic distractor design similarly implements a form of contrastive learning by organising examples that emphasise critical distinctions between alternations. This enables models to learn from the structured contrasts between valid and invalid linguistic transformations.

\paragraph{Linguistic Rule Learning in Neural Networks}
Neural approaches to linguistic rule learning are challenging despite recent advances. Neural language models demonstrate emergent syntactic capabilities~\citep{linzen-etal-2016-assessing, gulordava-etal-2018-colorless, hewitt-manning-2019-structural,mueller-etal-2024-context} but require large-scale training. \citet{kann-etal-2019-verb} examine alternations in transformer embeddings with limited success. \citet{thrush2020investigating} find that systematic selectional preferences require careful training design. LLMs also struggle on metalinguistic tasks~\citep{thrush-etal-2024-strange}.
\citet{merlo-2023-blackbird} introduces Blackbird Language Matrices (BLM) as structured evaluation paradigms for systematic rule assessment, enabling controlled testing of linguistic competence. However, they show limited efficiency in data-lean settings.
Our work addresses these efficiency challenges through analogical paradigm organisation, building on the BLM framework to achieve competitive argument structure learning with minimal training examples while demonstrating robustness across alternation types.

\section{Conclusion}

We investigated whether computational approaches inspired by cognitive principles can enable sample-efficient linguistic rule learning.
Our analogical paradigm organisation approach operationalises three organisational strategies, analogical structure, contrastive learning, and minimal contextual cues, through systematic input structuring rather than architectural scaling.

Our results demonstrate substantial sample efficiency gains across multiple evaluation dimensions. Lightweight models (BERT+CNN, $\sim0.5M$ parameters) trained on only $100$ structured examples achieved $F1=0.95$, and, in this setting, reached higher $F1$ than zero-shot \texttt{GPT-o3} ($F1=0.87$) on the same structured tasks, and stabilising performance with just $1 000-1 200$ training instances. Component analysis confirmed that each organisational principle contributes systematically, with analogical structure showing the largest effect. Cross-phenomenon validation with \textit{bake}-class verbs and cross-type generalisation experiments confirmed that benefits extend beyond construction-specific optimisations.

These findings have practical and theoretical implications. 
Practically, our approach enables competitive linguistic rule learning with much fewer task-specific examples than conventional LLM-based setups in this controlled setting.
Theoretically, we demonstrate that systematic input organisation can achieve efficiency gains that complement rather than require architectural scaling. This suggests that cognitive-inspired data structuring is a distinct optimisation dimension for linguistic rule induction.

Future work should explore cross-linguistic validation, integration with pretraining objectives, and automated structure discovery methods. 
Overall, our study shows that computational approaches inspired by cognitive principles can support sample-efficient linguistic rule learning and offer a complementary perspective on the behaviour of lightweight models and prompted LLMs.

\newpage

\section{Limitations}

While our approach demonstrates efficiency gains, it is not without limitations.

 \paragraph{Expert Dependency} Our structured paradigms require expert linguistic knowledge to design appropriate analogical mappings and systematic distractors. In practice, we follow a hybrid process where experts select and validate seed sentences and design a small set of templates, and full paradigms are then generated automatically. Although we demonstrate cross-phenomenon generalisability with \textit{bake}-class verbs, scaling this approach to diverse linguistic phenomena would require either linguistic expertise or automated methods for discovering optimal organisational structures.

 \paragraph{Pre-training Attribution}
 Our lightweight models rely on pre-trained encoders (BERT, RoBERTa, ELECTRA) that already encode linguistic knowledge from unstructured data. This creates attribution challenges in isolating whether performance gains stem purely from our organisational principles or from interactions with pre-existing linguistic representations. While our \textsc{Base} vs. \textsc{Shuffled} comparisons control for this confound, the fundamental attribution question remains.

 \paragraph{Language and Phenomenon Scope}
 Our evaluation was exclusively on English verb alternations, a specific subset of linguistic phenomena at the syntax-semantics interface. Cross-linguistic validation and extension to other grammatical constructions (e.g., morphological patterns, syntactic transformations) are needed to establish broader applicability of our organisational principles.

\paragraph{Computational Comparison Framework}
Our comparison between trained lightweight models and zero-shot and few-shot LLMs places the models in different learning regimes. The lightweight models are fine-tuned on $100-1 000$ task-specific examples. The LLMs are evaluated in zero-shot and few-shot prompt-based settings without parameter updates.
These settings are not matched in computational cost or data exposure. Our claims about lightweight models and LLMs are therefore restricted to this contrast and do not extend to general conclusions about LLM capabilities under full fine-tuning or matched training budgets.

\paragraph{Task Specificity}
Our approach targets pattern completion tasks that may favour analogical reasoning. This controlled design allows us to isolate the effects of input organisation on linguistic rule induction, but it does not cover broad downstream applications. Extension to diverse NLP applications (e.g., generation, open-ended reasoning) would provide stronger evidence for the general usability of our organisational principles.

\section*{Ethics}
Our research presents minimal ethical concerns. 
Our synthetically generated data focuses on grammatical verb alternations rather than semantic content, minimising bias propagation risks from LLM training data.

\section*{Acknowledgements}
We thank the anonymous reviewers of ACL Rolling Review and EACL for their constructive feedback. We also thank colleagues, Vivi Nastase and Giuseppe Samo at Idiap Research Institute for valuable discussions. We gratefully acknowledge the partial support of this work by the Swiss National Science Foundation, through grant TMAG-1\_209426 to PM.

\bibliography{custom}

\appendix
\label{sec:appendix}

 \section{Linguistic Phenomena}
\label{sec:app:ling_phenomenon}

\subsection{Causative/Inchoative Alternation (\textit{roll}-class verbs)}

The class of \textit{roll} verbs, as categorised by \citet{Levin93}, comprises verbs expressing dynamic actions with inherent motion characteristics. This class shows systematic alternation between:

\begin{itemize} \item \textbf{Intransitive}: Theme as subject (e.g., ``\textit{The ball rolled}'') \item \textbf{Transitive}: Agent causes Theme motion (e.g.,``\textit{The player rolled the ball}'') \end{itemize}

This syntactic-semantic interface represents a complex mapping challenge for computational models.

\paragraph{Verb Inventory} The verbs within this class include: \textit{bounce}, \textit{coil}, \textit{drift}, \textit{drop}, \textit{float}, \textit{glide}, \textit{move}, \textit{revolve}, \textit{roll}, \textit{rotate}, \textit{slide}, \textit{spin}, \textit{swing}, \textit{turn}, \textit{twirl}, \textit{twist}, \textit{whirl}, and \textit{wind}.

\subsection{Cross-Phenomenon Validation (\textit{bake}-class verbs)}

Unspecified object alternation with \textit{bake}-class verbs provides syntactically similar but semantically distinct validation:

\begin{itemize} \item \textbf{Transitive}: ``\textit{The chef baked a cake}'' \item \textbf{Intransitive}:``\textit{The chef baked}'' (object understood) \end{itemize}

\paragraph{Verb Inventory} The verbs within this class include: \textit{bake}, \textit{carve}, \textit{chop}, \textit{clean}, \textit{cook}, \textit{crochet}, \textit{draw}, \textit{drink}, \textit{dust}, \textit{eat}, \textit{embroider}, \textit{hum}, \textit{hunt}, \textit{fish}, \textit{iron}, \textit{knead}, \textit{knit}, \textit{mend}, \textit{milk}, \textit{mow}, \textit{nurse}, \textit{pack}, \textit{paint}, \textit{play}, \textit{plow}, \textit{polish}, \textit{read}, \textit{recite}, \textit{sew}, \textit{sculpt}, \textit{sing}, \textit{sketch}, \textit{sow}, \textit{study}, \textit{sweep}, \textit{teach}, \textit{type}, \textit{vacuum}, \textit{wash}, \textit{weave}, \textit{whittle} and \textit{write}.

This tests whether organisational benefits generalise across alternation types rather than being specific to a single alternation type.

\section{Experimental Setup Details}
\label{sec:app:setup}

\subsection{Ablation Design Formalisation} \label{sec:app:ablation}

We systematically manipulate context structures to isolate organisational components. Given a full context matrix $C_{\text{full}} \in \mathbb{R}^{m \times n}$ and transformation matrix $T \in \{0,1\}^{m \times n}$, transformed contexts are computed as: \[ C_{\text{trans}} = C_{\text{full}} \odot T \]

\paragraph{Transformation Matrices}

\begin{itemize}
    \item \textsc{NoAnalogy Context}
        \[
        T_{\text{af}} = \begin{bmatrix}
        0 & 0 & 0 & 0 \\
        1 & 1 & 1 & 0 
        \end{bmatrix}
        \]
 
    \item \textsc{NoSoftCue Context}
    \[
    T_{\text{scf}} = \begin{bmatrix}
    1 & 0 & 0 & 1 \\
    1 & 0 & 0 & 0 \\
    \end{bmatrix}
    \]
   
    \item For \textsc{Transposed Context}, it simply involves taking the transpose of the full context matrix:
    \[
C_{\text{trans}} = C_{\text{full}}^T
\]

\end{itemize}

\subsection{Statistical Analysis Framework}

\paragraph{Multi-factor ANOVA Design}
We use a five-way ANOVA with factors:
Structure (5 levels), Model (8 levels), Shots (3 levels), CoT (2 levels), and Data Type (2 levels).

\paragraph{Generalisation Gap Metric}
We define the generalisation gap as
\begin{equation}
  \text{GenGap}(t,s)
  = \frac{1}{|M|}
    \sum_{m \in M}
    \bigl[ F1_{m,s}^{\text{seen}}(t) - F1_{m,s}^{\text{unseen}}(t) \bigr]
  \label{eq:gen-gap-app}
\end{equation}
where $t$ is the training size, $s$ the structure type, $M$ the model set, and smaller gaps indicate better generalisation. Smoothing uses a centred rolling window ($size=3$) with confidence bands representing standard error propagation.

\section{Model Implementation}
\label{sec:app:model_spec}

\subsection{Encoder Comparison}

\begin{figure}[ht!] \centering \includegraphics[width=\linewidth]{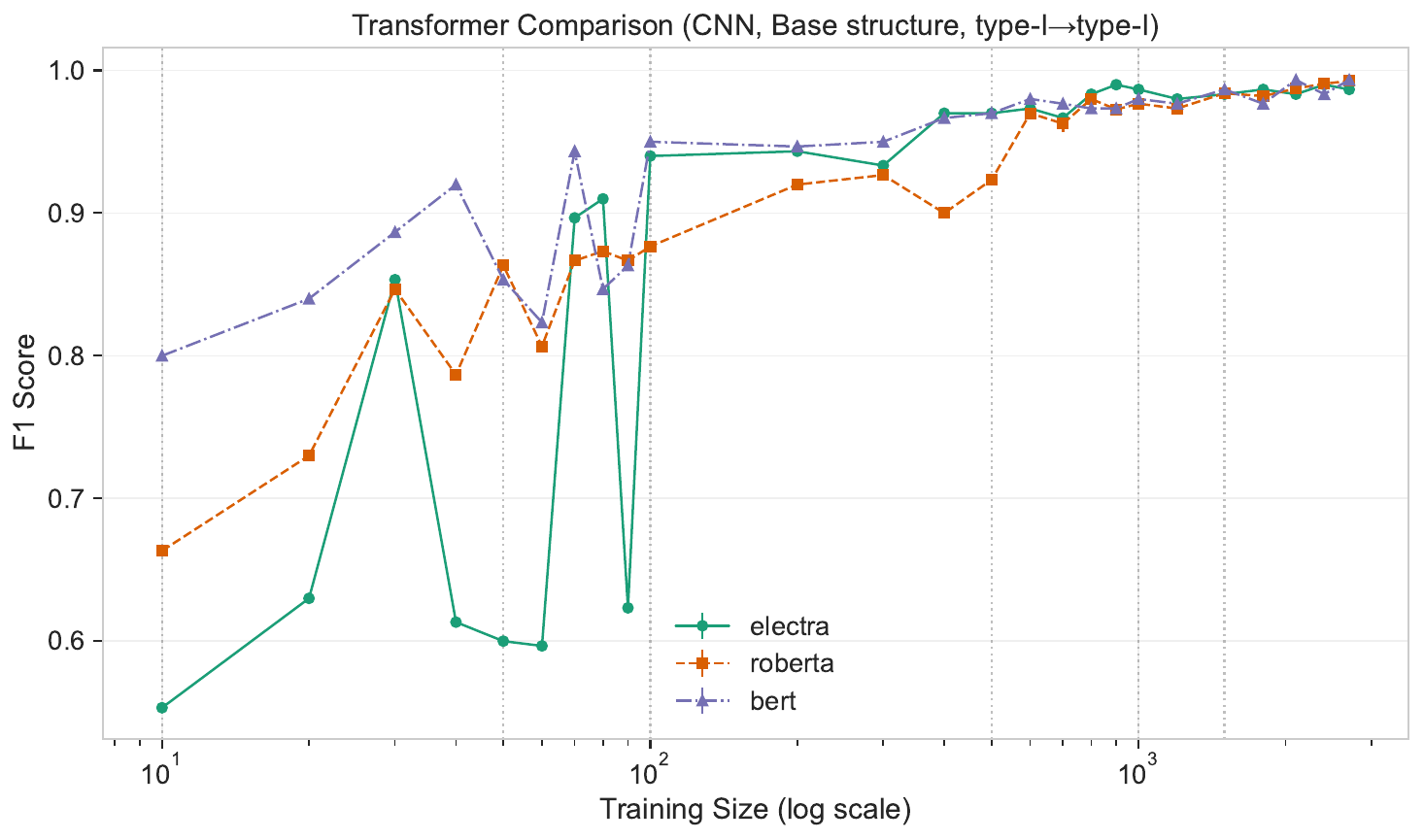} \caption{F1 performance as a function of training size, comparing encoder architectures (BERT, RoBERTa, ELECTRA) with the CNN architecture on \textsc{Base} structure. Train and test on \textit{Type I} data.} \label{fig:f-app-transformer} \end{figure}

We employ BERT (\texttt{bert-base-multilingual-cased}) as our primary encoder to emphasise contributions attributable to data organisation rather than encoder capabilities, as evidenced by Figure~\ref{fig:f-app-transformer}.

\subsection{Architecture Specifications}

\paragraph{Convolutional Neural Network (CNN)} 
\begin{itemize} 
\item Input: array of embeddings, size $7 \times 768$ 
\item Convolution layers: three 2D convolutional layers
    \begin{itemize} 
    \item kernel size: $3 \times 3$, Stride: 1, No dilation \end{itemize} 
\item Output: fully connected layer compressing to size $768$ \item Function: localises sequential patterns for analogical mapping 
\end{itemize}

\paragraph{Feed-Forward Neural Network (FFNN)} 
\begin{itemize} 
\item Input: concatenated sentence embeddings, size $7 \times 768$ 
\item Architecture: three fully connected layers 
\item Compression: $7 \times 768 \rightarrow 3.5 \times 768 \rightarrow 3.5 \times 768 \rightarrow 768$ 
\item Function: distributed pattern integration across the entire context \end{itemize}

\subsection{Training Configuration}

Models are trained over $120$ epochs with learning rate $0.001$ using the Adam optimiser and batch size $100$. Early stopping with patience $10$ prevents overfitting. All models employ a max-margin objective using cosine similarity:

\begin{equation} \mathcal{L} = \sum_{\mathbf{e}_i \in \mathcal{A}} \max\left(0, 1 + \cos(\mathbf{e}_i, \mathbf{e}_{\text{pred}})-\cos(\mathbf{e}_c, \mathbf{e}_{\text{pred}})\right) \end{equation}

where $\mathbf{e}_c$ is the correct answer embedding, $\mathbf{e}_i$ are incorrect distractor embeddings, and $\mathbf{e}_{\text{pred}}$ is the predicted completion. Models are run across three independent trials with seed $42$ to ensure statistical reliability.

\subsection{Computational Environment}

\paragraph{Hardware Configuration} \begin{itemize} \item \textbf{Workstation:} HP PAIR Workstation Z4 G4 MIT \item \textbf{Processor:} Intel Xeon W-2255 \item \textbf{RAM:} 64 GB \item \textbf{GPU:} MSI GeForce RTX 3090 VENTUS 3X OC, 24 GB GDDR6X \end{itemize}

\paragraph{Software Environment} \begin{itemize} \item Ubuntu: 22.04, Python: 3.11.5, CUDA: 11.8.0 \item PyTorch: 2.2.2, Transformers: 4.39.1, scikit-learn: 1.5.0 \item NumPy: 1.25.0, pandas: 1.4.3 \end{itemize}

\section{Complementary Experimental Results}
\label{sec:app:results_comp}

\subsection{Large Language Model Specifications}
\label{sec:app:llm_config}
We evaluate eight advanced models across four families with varying parameter scales. This systematic comparison allows us to assess how organisational structure affects different model capacities and architectural approaches.

\begin{table*}[ht!]
\centering
\small
\setlength{\tabcolsep}{4pt}
\begin{tabular}{llllll}
\hline
\textbf{Model Family} & \textbf{Variant} & \textbf{Parameters} & \textbf{Context} & \textbf{Type} & \textbf{Release} \\ \hline
\multicolumn{6}{c}{\textit{Reasoning Models}}                                \\ \hline
DeepSeek & deepseek-R1-0528       & ~670B       & 64K  & Reasoning & 2025-05 \\
OpenAI   & gpt-o3                 & Undisclosed & 128K & Reasoning & 2025-04 \\
OpenAI   & gpt-o3-mini            & Undisclosed & 128K & Reasoning & 2025-01 \\
Alibaba  & qwq-32B                & 32B         & 32K  & Reasoning & 2025-03 \\ \hline
\multicolumn{6}{c}{\textit{Standard Models}}                                 \\ \hline
DeepSeek & deepseek-V3-0324       & 671B        & 64K  & Standard  & 2025-03 \\
Meta     & llama-3.3-70B-Instruct & 70B         & 128K & Standard  & 2024-12 \\
Meta     & llama-3.2-3B-Instruct  & 3B          & 128K & Standard  & 2024-09 \\
Alibaba  & qwen3-32B              & 32B         & 128K & Standard  & 2025-04 \\ \hline
\end{tabular}

\caption{Large language models evaluated in experiments. Context window indicates maximum input sequence length in tokens.}
\label{tab:llm}
\end{table*}

\subsection{Statistical Analysis Results}
\label{sec:app:llm_anova}
Complete multi-factor ANOVA results demonstrating significant main effects for structure, model, and number of shots across all experimental conditions.

\begin{table*}[]
\centering
\begin{tabular}{l|ll}
\hline
Factors                                            & F         & p          \\ \hline
C(Structure)                                       & 9284.541  & $p<0.05$ \\
C(Model)                                           & 9581.462  & $p<0.05$ \\
C(Num\_Shots)                                      & 14125.142 & $p<0.05$ \\
C(CoT)                                             & 9.562     & $p<0.05$ \\
C(Data\_Type)                                      & 191.786   & $p<0.05$ \\
C(Structure):C(Model)                              & 313.487   & $p<0.05$ \\
C(Structure):C(Num\_Shots)                         & 170.116   & $p<0.05$ \\
C(Model):C(Num\_Shots)                             & 664.883   & $p<0.05$ \\
C(Structure):C(CoT)                                & 2.125     & 0.0601     \\
C(Model):C(CoT)                                    & 3.625     & $p<0.05$ \\
C(Num\_Shots):C(CoT)                               & 5.399     & $p<0.05$ \\
C(Structure):C(Data\_Type)                         & 115.123   & $p<0.05$ \\
C(Model):C(Data\_Type)                             & 37.116    & $p<0.05$ \\
C(Num\_Shots):C(Data\_Type)                        & 34.928    & $p<0.05$ \\
C(CoT):C(Data\_Type)                               & 0.078     & 0.7795     \\
C(Structure):C(Model):C(Num\_Shots)                & 55.286    & $p<0.05$ \\
C(Structure):C(Model):C(CoT)                       & 1.644     & $p<0.05$ \\
C(Structure):C(Num\_Shots):C(CoT)                  & 2.265     & $p<0.05$ \\
C(Model):C(Num\_Shots):C(CoT)                      & 1.889     & $p<0.05$ \\
C(Structure):C(Model):C(Data\_Type)                & 13.123    & $p<0.05$ \\
C(Structure):C(Num\_Shots):C(Data\_Type)           & 17.556    & $p<0.05$ \\
C(Model):C(Num\_Shots):C(Data\_Type)               & 11.177    & $p<0.05$ \\
C(Structure):C(CoT):C(Data\_Type)                  & 0.368     & 0.8709     \\
C(Model):C(CoT):C(Data\_Type)                      & 0.214     & 0.9824     \\
C(Num\_Shots):C(CoT):C(Data\_Type)                 & 6.551     & $p<0.05$ \\
C(Structure):C(Model):C(Num\_Shots):C(CoT)         & 0.899     & 0.7079     \\
C(Structure):C(Model):C(Num\_Shots):C(Data\_Type)  & 7.100     & $p<0.05$ \\
C(Structure):C(Model):C(CoT):C(Data\_Type)         & 0.919     & 0.6045     \\
C(Structure):C(Num\_Shots):C(CoT):C(Data\_Type)    & 1.748     & 0.0657     \\
C(Model):C(Num\_Shots):C(CoT):C(Data\_Type)        & 1.221     & 0.2531     \\
C(Structure):C(Model):C(Num\_Shots):C(CoT):C(Da... & 0.944     & 0.6086     \\ \hline
\end{tabular}
\caption{Multi-ANOVA tests on data (Structure, Data\_Type) and experimental (Model, Num\_Shots, CoT) factors.}
\label{tab:multi-anova-llm}
\end{table*}

\subsection{Comprehensive Performance Results}
\label{sec:app:llm_results}
Complete results across all models, structures, data types, and number of shot configurations with and without Chain-of-Thought reasoning.

\begin{table*}[t]
\small
\centering
\setlength{\tabcolsep}{4pt}
\begin{tabular}{llllll}
\toprule
                           Model Config &          Base &     Noanalogy &     Nosoftcue &      Shuffled &     Transpose \\
\midrule
            deepseek-R1 (0-shot, w-CoT) & 0.743 ± 0.048 & 0.692 ± 0.037 & 0.711 ± 0.049 & 0.614 ± 0.043 & 0.655 ± 0.050 \\
           deepseek-R1 (0-shot, wo-CoT) & 0.719 ± 0.063 & 0.707 ± 0.018 & 0.720 ± 0.020 & 0.641 ± 0.030 & 0.681 ± 0.035 \\
            deepseek-R1 (1-shot, w-CoT) & 0.857 ± 0.004 & 0.876 ± 0.048 & 0.802 ± 0.017 & 0.788 ± 0.012 & 0.851 ± 0.028 \\
           deepseek-R1 (1-shot, wo-CoT) & 0.848 ± 0.009 & 0.884 ± 0.054 & 0.802 ± 0.005 & 0.765 ± 0.021 & 0.873 ± 0.071 \\
            deepseek-R1 (5-shot, w-CoT) & 0.884 ± 0.052 & 0.994 ± 0.004 & 0.875 ± 0.072 & 0.871 ± 0.073 & 0.927 ± 0.076 \\
           deepseek-R1 (5-shot, wo-CoT) & 0.883 ± 0.058 & 0.911 ± 0.065 & 0.866 ± 0.069 & 0.871 ± 0.073 & 0.859 ± 0.007 \\
            deepseek-V3 (0-shot, w-CoT) & 0.537 ± 0.193 & 0.859 ± 0.089 & 0.397 ± 0.163 & 0.445 ± 0.107 & 0.196 ± 0.130 \\
           deepseek-V3 (0-shot, wo-CoT) & 0.487 ± 0.409 & 0.912 ± 0.102 & 0.630 ± 0.320 & 0.483 ± 0.410 & 0.139 ± 0.177 \\
            deepseek-V3 (1-shot, w-CoT) & 0.729 ± 0.015 & 0.961 ± 0.008 & 0.634 ± 0.019 & 0.620 ± 0.056 & 0.544 ± 0.082 \\
           deepseek-V3 (1-shot, wo-CoT) & 0.725 ± 0.011 & 0.963 ± 0.008 & 0.632 ± 0.019 & 0.604 ± 0.033 & 0.509 ± 0.092 \\
            deepseek-V3 (5-shot, w-CoT) & 0.895 ± 0.063 & 0.994 ± 0.002 & 0.853 ± 0.058 & 0.703 ± 0.021 & 0.828 ± 0.100 \\
           deepseek-V3 (5-shot, wo-CoT) & 0.963 ± 0.007 & 0.995 ± 0.003 & 0.899 ± 0.013 & 0.716 ± 0.036 & 0.867 ± 0.076 \\
                 gpt-o3 (0-shot, w-CoT) & 0.852 ± 0.008 & 0.754 ± 0.031 & 0.844 ± 0.053 & 0.688 ± 0.041 & 0.838 ± 0.012 \\
                gpt-o3 (0-shot, wo-CoT) & 0.884 ± 0.057 & 0.754 ± 0.022 & 0.840 ± 0.064 & 0.700 ± 0.030 & 0.834 ± 0.015 \\
                 gpt-o3 (1-shot, w-CoT) & 0.891 ± 0.087 & 0.995 ± 0.012 & 0.858 ± 0.076 & 0.783 ± 0.053 & 0.869 ± 0.066 \\
                gpt-o3 (1-shot, wo-CoT) & 0.911 ± 0.082 & 0.990 ± 0.008 & 0.834 ± 0.031 & 0.785 ± 0.046 & 0.894 ± 0.079 \\
                 gpt-o3 (5-shot, w-CoT) & 0.939 ± 0.072 & 0.998 ± 0.004 & 0.884 ± 0.094 & 0.858 ± 0.075 & 0.922 ± 0.086 \\
                gpt-o3 (5-shot, wo-CoT) & 0.925 ± 0.078 & 1.000 ± 0.000 & 0.842 ± 0.019 & 0.830 ± 0.033 & 0.938 ± 0.085 \\
            gpt-o3-mini (0-shot, w-CoT) & 0.754 ± 0.026 & 0.629 ± 0.028 & 0.606 ± 0.044 & 0.431 ± 0.041 & 0.641 ± 0.034 \\
           gpt-o3-mini (0-shot, wo-CoT) & 0.745 ± 0.047 & 0.644 ± 0.070 & 0.636 ± 0.052 & 0.410 ± 0.014 & 0.573 ± 0.101 \\
            gpt-o3-mini (1-shot, w-CoT) & 0.809 ± 0.015 & 0.863 ± 0.055 & 0.673 ± 0.024 & 0.542 ± 0.060 & 0.701 ± 0.063 \\
           gpt-o3-mini (1-shot, wo-CoT) & 0.812 ± 0.023 & 0.866 ± 0.059 & 0.647 ± 0.068 & 0.551 ± 0.044 & 0.730 ± 0.068 \\
            gpt-o3-mini (5-shot, w-CoT) & 0.894 ± 0.076 & 0.989 ± 0.013 & 0.774 ± 0.042 & 0.638 ± 0.044 & 0.767 ± 0.053 \\
           gpt-o3-mini (5-shot, wo-CoT) & 0.881 ± 0.086 & 0.995 ± 0.005 & 0.764 ± 0.045 & 0.622 ± 0.053 & 0.786 ± 0.047 \\
  llama-3.2-3B-Instruct (0-shot, w-CoT) & 0.093 ± 0.035 & 0.478 ± 0.033 & 0.109 ± 0.014 & 0.118 ± 0.015 & 0.061 ± 0.033 \\
 llama-3.2-3B-Instruct (0-shot, wo-CoT) & 0.105 ± 0.042 & 0.501 ± 0.034 & 0.112 ± 0.018 & 0.113 ± 0.015 & 0.066 ± 0.037 \\
  llama-3.2-3B-Instruct (1-shot, w-CoT) & 0.524 ± 0.079 & 0.761 ± 0.031 & 0.388 ± 0.044 & 0.279 ± 0.053 & 0.283 ± 0.049 \\
 llama-3.2-3B-Instruct (1-shot, wo-CoT) & 0.532 ± 0.055 & 0.757 ± 0.032 & 0.374 ± 0.039 & 0.271 ± 0.057 & 0.269 ± 0.036 \\
  llama-3.2-3B-Instruct (5-shot, w-CoT) & 0.456 ± 0.071 & 0.882 ± 0.045 & 0.330 ± 0.051 & 0.317 ± 0.043 & 0.308 ± 0.031 \\
 llama-3.2-3B-Instruct (5-shot, wo-CoT) & 0.491 ± 0.063 & 0.890 ± 0.038 & 0.313 ± 0.041 & 0.340 ± 0.051 & 0.319 ± 0.026 \\
 llama-3.3-70B-Instruct (0-shot, w-CoT) & 0.304 ± 0.083 & 0.700 ± 0.015 & 0.340 ± 0.034 & 0.337 ± 0.026 & 0.238 ± 0.110 \\
llama-3.3-70B-Instruct (0-shot, wo-CoT) & 0.298 ± 0.087 & 0.687 ± 0.017 & 0.309 ± 0.021 & 0.318 ± 0.018 & 0.223 ± 0.106 \\
 llama-3.3-70B-Instruct (1-shot, w-CoT) & 0.774 ± 0.021 & 0.845 ± 0.046 & 0.515 ± 0.036 & 0.491 ± 0.042 & 0.592 ± 0.073 \\
llama-3.3-70B-Instruct (1-shot, wo-CoT) & 0.783 ± 0.047 & 0.783 ± 0.080 & 0.499 ± 0.015 & 0.447 ± 0.046 & 0.594 ± 0.066 \\
 llama-3.3-70B-Instruct (5-shot, w-CoT) & 0.964 ± 0.055 & 0.996 ± 0.001 & 0.875 ± 0.084 & 0.737 ± 0.035 & 0.952 ± 0.045 \\
llama-3.3-70B-Instruct (5-shot, wo-CoT) & 0.942 ± 0.071 & 0.976 ± 0.051 & 0.870 ± 0.091 & 0.734 ± 0.021 & 0.931 ± 0.056 \\
              qwen3-32B (0-shot, w-CoT) & 0.786 ± 0.122 & 0.596 ± 0.222 & 0.736 ± 0.029 & 0.441 ± 0.009 & 0.538 ± 0.074 \\
             qwen3-32B (0-shot, wo-CoT) & 0.659 ± 0.088 & 0.662 ± 0.059 & 0.752 ± 0.069 & 0.429 ± 0.016 & 0.535 ± 0.065 \\
              qwen3-32B (1-shot, w-CoT) & 0.808 ± 0.011 & 0.797 ± 0.018 & 0.792 ± 0.018 & 0.532 ± 0.014 & 0.729 ± 0.046 \\
             qwen3-32B (1-shot, wo-CoT) & 0.811 ± 0.021 & 0.788 ± 0.009 & 0.800 ± 0.012 & 0.525 ± 0.028 & 0.740 ± 0.046 \\
              qwen3-32B (5-shot, w-CoT) & 0.839 ± 0.003 & 0.896 ± 0.070 & 0.830 ± 0.017 & 0.636 ± 0.049 & 0.815 ± 0.007 \\
             qwen3-32B (5-shot, wo-CoT) & 0.833 ± 0.009 & 0.917 ± 0.067 & 0.842 ± 0.063 & 0.619 ± 0.021 & 0.808 ± 0.010 \\
                qwq-32B (0-shot, w-CoT) & 0.776 ± 0.035 & 0.724 ± 0.023 & 0.747 ± 0.026 & 0.504 ± 0.018 & 0.672 ± 0.076 \\
               qwq-32B (0-shot, wo-CoT) & 0.752 ± 0.060 & 0.732 ± 0.011 & 0.731 ± 0.036 & 0.462 ± 0.042 & 0.652 ± 0.076 \\
                qwq-32B (1-shot, w-CoT) & 0.843 ± 0.010 & 0.821 ± 0.015 & 0.793 ± 0.007 & 0.598 ± 0.042 & 0.762 ± 0.056 \\
               qwq-32B (1-shot, wo-CoT) & 0.843 ± 0.009 & 0.822 ± 0.015 & 0.792 ± 0.011 & 0.600 ± 0.047 & 0.755 ± 0.048 \\
                qwq-32B (5-shot, w-CoT) & 0.863 ± 0.006 & 0.863 ± 0.004 & 0.831 ± 0.017 & 0.691 ± 0.021 & 0.846 ± 0.021 \\
               qwq-32B (5-shot, wo-CoT) & 0.864 ± 0.002 & 0.863 ± 0.003 & 0.829 ± 0.009 & 0.680 ± 0.017 & 0.836 ± 0.027 \\
\bottomrule
\end{tabular}
\caption{Performance (F1 scores in \% $\pm$ standard deviation) of all LLMs validated across all configurations. }
\label{tab:llm_all}
\end{table*}

\end{document}